\documentclass[10pt,twocolumn,letterpaper]{article}

\usepackage{cvpr}
\usepackage{graphicx}
\usepackage{amsmath}
\usepackage{amssymb}
\usepackage{booktabs}

\usepackage{color}

\usepackage{epstopdf}
\usepackage{multirow}
\usepackage{amsmath,amssymb}
\usepackage{algorithm}
\usepackage{algorithmicx}
\usepackage{algpseudocode}

\usepackage{booktabs}
\usepackage{makecell}
\usepackage{textcomp}
\usepackage{mathrsfs}
\usepackage{multicol}
\usepackage{mdframed}

\usepackage[pagebackref,breaklinks,colorlinks]{hyperref}

\usepackage[capitalize]{cleveref}
\crefname{section}{Sec.}{Secs.}
\Crefname{section}{Section}{Sections}
\Crefname{table}{Table}{Tables}
\crefname{table}{Tab.}{Tabs.}

\begin{document}

\title{EchoMimic: Lifelike Audio-Driven Portrait Animations through Editable Landmark Conditions}

\author{Zhiyuan Chen\thanks{Equal contribution}  \\
Ant Group\\
{\tt\small juzhen.czy@antgroup.com}
 \and
Jiajiong Cao$^*$  \\
Ant Group\\
{\tt\small jiajiong.caojiajio@antgroup.com}
 \and
Zhiquan Chen  \\
Ant Group\\
{\tt\small zhiquan.zhiquanche@antgroup.com}
\and
Yuming Li  \\
Ant Group\\
{\tt\small luoque.lym@antgroup.com}
% \and
% Yuming Li  \\
% Ant Group\\
% {\tt\small luoque.lym@antgroup.com}
 \and
Chenguang Ma  \\
Ant Group\\
{\tt\small chenguang.mcg@antgroup.com} \\
% \and
}
% \author{\href{https://badtobest.github.io/echomimic.html}{Project Page}}
\maketitle
% \begin{multicols}{2}
% % Project page link
% \begin{center}
% \url{http://www.example-project-page.com}
% \end{center}
% \end{multicols}

% \begin{@twocolumntrue}
% \Needspace{10\baselineskip}
% \begin{onecolumn}
% \textbf{Project Page:} \url{https://badtobest.github.io/echomimic.html}
% \end{onecolumn}
% \footnotetext{$^\dagger$ Equal contribution.}

\begin{quote}
\textbf{Project Page:} \url{https://badtobest.github.io/echomimic.html}
\end{quote}

% \begin{multicols}{2}
% \end{@twocolumntrue}

% \begin{multicols}{2}
% \lipsum[1] % 生成第一段示例文本

% \columnbreak % 确保下一个环境开始于新的列

% \lipsum[2] % 生成第二段示例文本
% \end{multicols}

% \begin{quote}
% \raggedright % 可选，使引用的文本两端对齐
% 这是一个横跨两栏的引用环境示例。在这个引用环境中，文本会自动流过整个页面宽度，而不是限制在单个列内。
% \end{quote}

% \begin{multicols}{2}
% \lipsum[3] % 生成第三段示例文本
% \end{multicols}

%%%%%%%%% ABSTRACT
\begin{abstract}
The area of portrait image animation, propelled by audio input, has witnessed notable progress in the generation of lifelike and dynamic portraits. Conventional methods are limited to utilizing either audios or facial key points to drive images into videos, while they can yield satisfactory results, certain issues exist. For instance, methods driven solely by audios can be unstable at times due to the relatively weaker audio signal, while methods driven exclusively by facial key points, although more stable in driving, can result in unnatural outcomes due to the excessive control of key point information. In addressing the previously mentioned challenges, in this paper, we introduce a novel approach which we named EchoMimic. EchoMimic is concurrently trained using both audios and facial landmarks. Through the implementation of a novel training strategy, EchoMimic is capable of generating portrait videos not only by audios and facial landmarks individually, but also by a combination of both audios and selected facial landmarks. EchoMimic has been comprehensively compared with alternative algorithms across various public datasets and our collected dataset, showcasing superior performance in both quantitative and qualitative evaluations. Additional visualization and access to the source code can be located on the EchoMimic project page.
\end{abstract}

%%%%%%%%% BODY TEXT
\section{Introduction} \label{sec:intro}

The recent advancement in image generation has been greatly advanced by the introduction and effectiveness of Diffusion Models \cite{rombach2022high,dhariwal2021diffusion,ho2020denoising}. Through rigorous training on large image datasets and a stepwise generation process, these models enable the creation of hyper-realistic images with unprecedented detail. This innovative progress has not only reshaped the field of generative models but has also expanded its application into video synthesis for crafting vivid and engaging visual narratives. In the realm of video synthesis, a significant focus lies in generating human-centric content, notably talking head animations, which involves translating audio inputs into corresponding facial expressions. This task is inherently complex due to the intricate nature and diversity of human facial movements. Conventional methods, despite simplifying the process through constraints such as 3D facial modeling or motion extraction from base videos, often compromise the richness and authenticity of facial expressions. 

\begin{figure}
  \centering
  \includegraphics[width=0.95\linewidth]{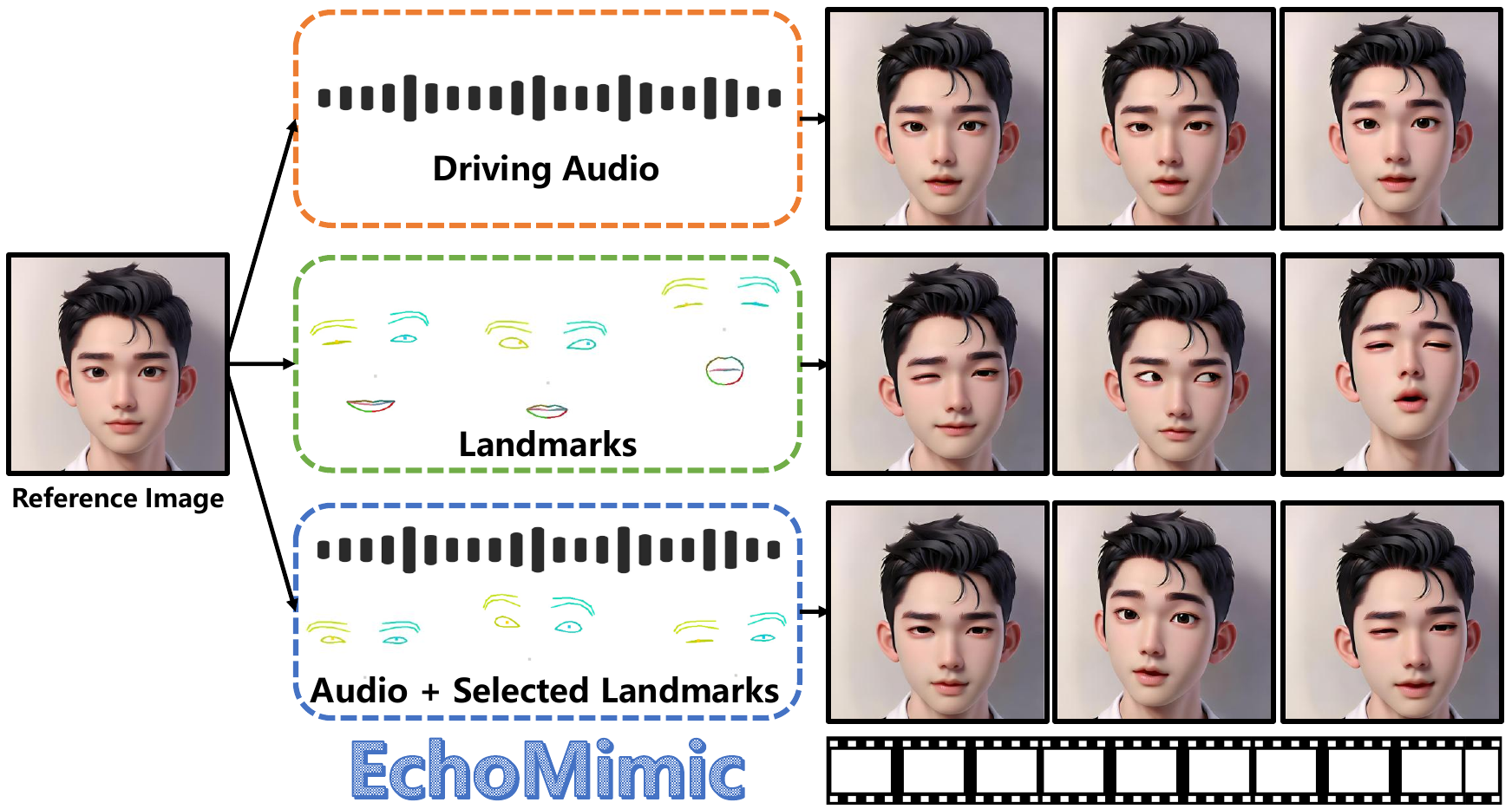}
  \caption{EchoMimic is capable of generating portrait videos by audios, facial landmarks and a combination of both audios and selected facial landmarks.}
  \label{fig:Figure1} %% label for entire figure
\end{figure}

Portrait animation, a subset of this domain, involves transferring motion and expressions from a source video to a target portrait image using Generative Adversarial Networks (GANs) and diffusion models. Despite the structured two-stage process followed by GAN-based methods \cite{drobyshev2022megaportraits,liu2023human}, which includes feature warping and refinement, they are limited by GAN performance and inaccurate motion depiction, resulting in unrealistic outputs. In contrast, diffusion models have exhibited superior generation capacity, leading to their adaptation for portrait animation tasks. Efforts to enhance these models with specialized modules have been pursued to preserve the portrait's identity and accurately model target expressions. However, challenges such as distortions and artifacts persist, particularly when working with unconventional portrait types, due to inadequate motion representation and inappropriate loss functions for the specific demands of portrait animation.
The field faces the dual challenges of synchronizing lip movements, facial expressions, and head poses with audio inputs, and producing visually appealing, high-fidelity animations with consistent temporal coherence. While parametric model-based solutions rely on audio-driven intermediate representations such as 3DMM \cite{sun2023vividtalk}, their limitations are imposed by the adequacy of these representations. Decoupled representation learning in the latent space provides an alternative approach by independently addressing the identity and non-identity aspects of facial features. However, it encounters difficulties in achieving comprehensive disentanglement and ensuring consistency across frames.

The quest for progress in portrait animation, particularly talking head image animation, holds substantial significance across various sectors, including gaming, media production, and education. Notable works such as Stable Diffusion (SD) and DiT (Diffusion Models with Transformers) \cite{peebles2023scalable} demonstrate notable advancements in this field. The incorporation of diffusion techniques and parametric or implicit representations of facial dynamics in a latent space facilitates the end-to-end generation of high-quality, realistic animations. However, traditional approaches are constrained to utilizing either audio or facial landmarks for driving images into videos. While these methods can produce satisfactory results, they are associated with specific limitations. For example, approaches driven solely by audio may experience instability due to the relatively weaker audio signal, whereas methods exclusively driven by facial landmarks, although more stable in driving, can lead to unnatural outcomes due to the excessive control of landmark information.

To address the aforementioned challenges, in this paper, we present a novel approach called EchoMimic. EchoMimic is concurrently trained using both audio signals and facial landmarks. Leveraging a novel training strategy, as shown in Fig. \ref{fig:Figure1}, EchoMimic demonstrates the ability to generate portrait videos using either audios or facial landmarks independently, as well as a combination of audios and selected facial landmarks. EchoMimic is extensively compared with alternative algorithms across diverse public datasets and our collected dataset, demonstrating superior performance in both quantitative and qualitative evaluations.

\begin{figure*}
  \centering
  \includegraphics[width=0.88\linewidth]{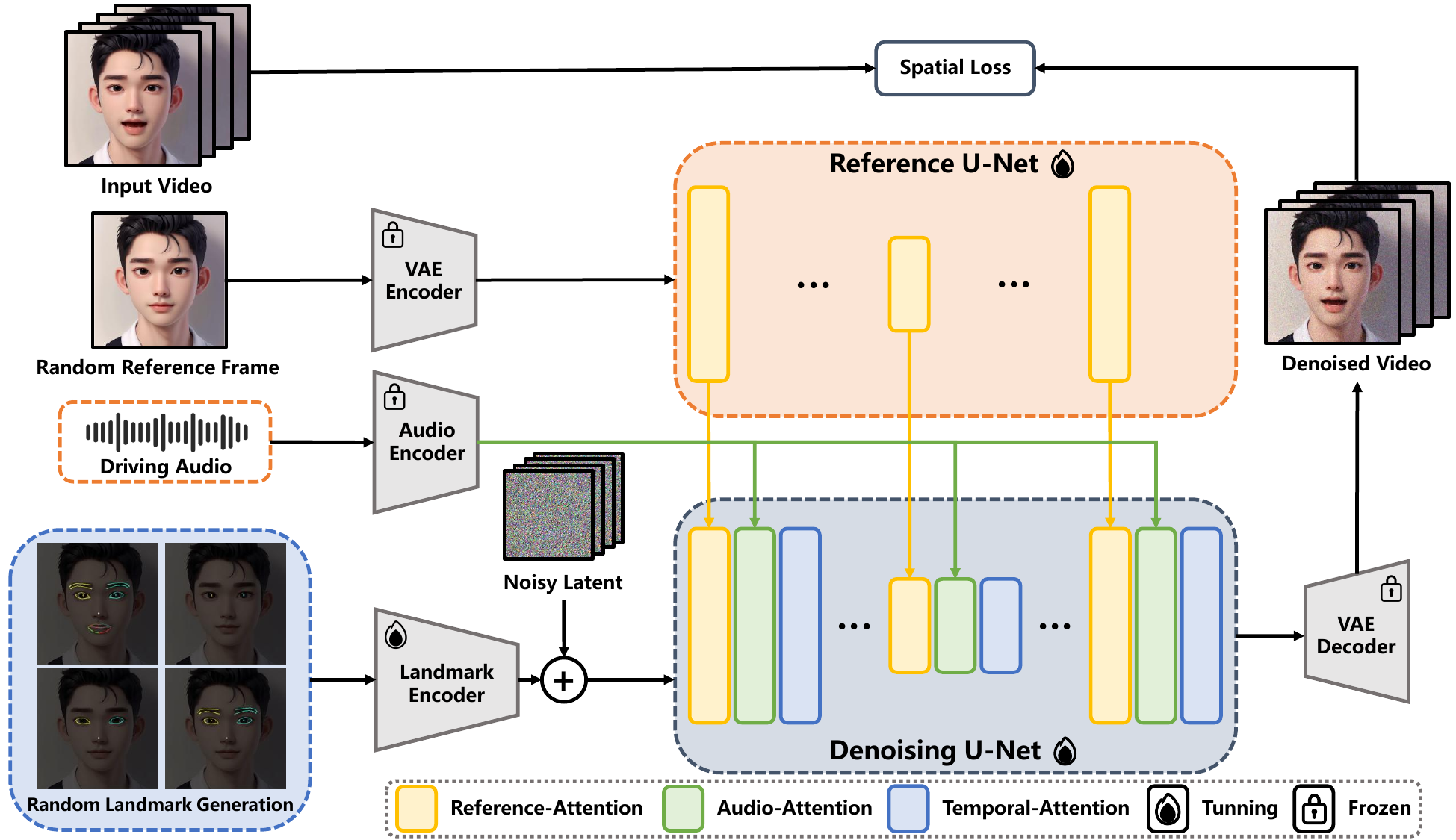}
  \caption{The overall pipeline of the proposed EchoMimic (EM) framework.}
  \label{fig:Figure1} %% label for entire figure
\end{figure*}

\section{Related Works}
\subsection{Diffusion Models}
Diffusion-based generative models have recently emerged as a cornerstone in computational creativity, demonstrating remarkable versatility and adaptability across a spectrum of multimedia tasks. These models have been effectively employed in the synthesis of novel imagery, the refinement and transformation of existing visuals, the production of dynamic video content, and the creation of intricate three-dimensional digital constructs.

A paradigmatic illustration of diffusion models' potential is exemplified by SD, which integrates a UNet framework to iteratively generate images conditioned on textual descriptions. This capability is honed through extensive training on large-scale multimodal datasets that meticulously link text to corresponding images. Once pre-trained, diffusion models exhibit extraordinary flexibility, facilitating a broad range of creative applications spanning both static and dynamic visual media. Innovative methodologies have sought to augment diffusion models by merging traditional UNet architectures with Transformer-based designs, equipped with temporal layers and three-dimensional convolutional capabilities, as seen in DiT. This amalgamation enables the effective training of models specifically tailored for text-conditioned video generation from scratch, resulting in superior outcomes in video synthesis. Moreover, diffusion models have gained traction for the generation of life-like animated portraits, commonly referred to as 'talking heads', showcasing their prowess in synthesizing realistic sequences of human-like speech animations.

\subsection{Portrait Animation: From Video to Image-Based Approaches}
The field of 'talking head' animation has evolved significantly, initially relying on video-based techniques for the synthesis of lip movements synchronized with audio inputs. Notably, Wav2Lip, a pioneering methodology, excels inoverlaying synthesized lip movements onto existing videocontent while ensuring audio-lip synchronicity through theemployment of a discriminator mechanism \cite{prajwal2020lip}. However, Wav2Lip's outputs can occasionally exhibit realism deficiencies, characterized by artifacts such as blurred visualsor distorted dental structures within the lower facial region.

Recent advancements have pivoted towards image-based methodologies, undergirded by diffusion models, to mitigatethe limitations of video-centric approaches and enhancethe realism and expressiveness of synthesized animations. Animate Anyone, a foundational work, has paved theway for expressive portrait video generation with minimalcontrol inputs \cite{hu2024animate}. Building upon this foundation, EMO\cite{tian2024emo} introduces a novel Frame Encoding module alongside robust control features, ensuring audio-driven animations maintain consistency across video frames, thereby bolstering the stability and naturalness of synthesized speech animations.

SadTalker\cite{zhang2023sadtalker} generates 3D motion coefficients (head pose, expression) of the 3D Morphable Model from audio and implicitly modulates a 3D-aware face render for talking head generation. To learn realistic motion coefficients, SadTalker explicitly model the connections between audio and different types of motion coefficients. The generated 3D motion coefficients are then mapped to the unsupervised 3D keypoints space of the proposed face render to synthesize the final video.

AniPortrait\cite{wei2024aniportrait} represents another significant stride in the domain, adeptly translating audio signals into detailed 3D facial structures before mapping them onto 2D faciallandmarks. The subsequent diffusion model, coupled with a motion module, renders these landmarks into temporally coherent video sequences, enriching the portrayal of natural emotions and enabling nuanced facial motion alterations and reenactments.

V-Express\cite{wang2024v}, expanding the horizons of audio-visual alignment, employs a layered structure to meticulously synchronize audio with the subtle dynamics of lip movements, facial expressions, and head poses. Its sophisticated facial loss function further refines the model's sensitivity to subtle emotional nuances and the overall aesthetic appeal of facial features.

Hallo\cite{xu2024hallo} contributes a hierarchical audio-driven visual synthesis approach for portrait image animation, addressing the intricacies of lip synchronization, expression, and pose alignment. By integrating diffusion-based generative models with a UNet denoiser and cross-attention mechanisms, it achieves enhanced control over expression diversity and pose variation, demonstrating improvements in video quality, lip sync precision, and motion variety.

However, despite the considerable strides made by image-based methods, several challenges remain. These methods typically condition synthesis on either audio or pose inputs separately, rarely integrating both concurrently. Furthermore, evaluation protocols often lean heavily on image-level metrics such as Fréchet Inception Distance (FID) and Expression Fréchet Inception Distance (E-FID), potentially overlooking critical aspects of facial structure and dynamics. Addressing these complexities is essential for advancing the realism and fidelity of 'talking head' animations, paving the way for more immersive and lifelike multimedia experiences. 

\section{Method}
\subsection{Preliminaries}\label{section:pre}
Our approach is grounded in Stable Diffusion (SD), a seminal framework in text-to-image (T2I) conversion that builds upon the Latent Diffusion Model (LDM)\cite{rombach2022high}. Central to SD is the application of a Variational Autoencoder (VAE)\cite{kingma2013auto}, which acts as an autoencoder. This mechanism transforms the original image's feature distribution, denoted as $x_{0}$, into a latent space representation $z_0$. The encoding phase captures the image essence as $z_0 = E(x_0)$, whereas the decoding counterpart reconstructs it back to $x_0 = D(z_0)$. This design significantly curtails computational expenses without compromising visual quality.

SD integrates principles from the Denoising Diffusion Probabilistic Model (DDPM)\cite{ho2020denoising} or its variant, the Denoising Diffusion Implicit Model (DDIM)\cite{song2020denoising}, introducing a strategic element of Gaussian noise $\epsilon$ to the latent representation $z_0$, yielding a temporally indexed noisy latent state $z_t$ at step $t$. The inferential phase of SD revolves around a dual objective: progressively eliminating this injected noise $\epsilon$ from $z_t$ and concurrently leveraging textual directives. By seamlessly incorporating text embeddings, SD directs the denoising process to yield images that adhere closely to the prescribed textual prompts, thereby realizing finely controlled, high-fidelity visual outputs. The objective function guiding the denoising process during training is formulated as follows:

\begin{equation}
\begin{aligned}
\label{equation:denoise}
\mathcal{L} = \mathbb{E}_{t,c,z_t,\epsilon}[||\epsilon-\epsilon_{\theta}(z_t,t,c)||^{2}]
\end{aligned}
\end{equation}

Here, $c$ signifies the text features extracted from the input prompt utilizing the CLIP\cite{radford2021learning} ViT-L/14 text encoder. Within the Stable Diffusion (SD) framework, the estimation of the noise $\epsilon$ is accomplished by a customized UNet\cite{ronneberger2015u} architecture. This UNet model has been augmented with a cross-attention mechanism, allowing for the effective integration of text features $c$ with the latent representation $z_t$, thereby enhancing the model's capability to generate images that are coherent with the provided text guidance.

\subsection{Model Architecture}\label{section:arch}
The foundational component of our proposed EchoMimic framework is the Denoising U-Net architecture, as depicted in Figure \ref{fig:Figure1}. In order to enhance the network's capacity to assimilate diverse inputs, EchoMimic integrates three specialized modules: Reference U-Net for encoding reference images, Landmark Encoder for guiding the network using facial landmarks, and Audio Encoder for encoding audio inputs. These modules collectively ensure a comprehensive and contextually rich encoding process, crucial for generating high-fidelity video content. Detailed descriptions of these architectures will be provided in the following sections.

\textbf{Denoising U-Net.} The Denoising U-Net is tailored to enhance multi-frame latent representations corrupted by noise across different conditions, drawing inspiration from the well-established SDv1.5 architecture and incorporating three distinct attention layers within each Transformer block. The initial Reference-Attention layer fosters adept encoding of the relationship between the current frame and reference images, while the second Audio-Attention layer captures the interaction between visual and audio content, operating on a spatial dimension. Additionally, the Temporal-Attention layer deploys a temporal-wise self-attention mechanism to decipher intricate temporal dynamics and relationships between consecutive video frames. These enhancements are pivotal for nuanced understanding and integration of spatial and temporal relationships across the network.

\textbf{Reference U-Net.} The reference image is crucial for preserving facial identity and background consistency within the EchoMimic framework. To facilitate this, we introduce the specialized module, the Reference U-Net, which mirrors the architectural design of SDv1.5 and operates in parallel with the Denoising U-Net. Within each Transformer block of the Reference U-Net, the self-attention mechanism is used to extract reference image features, subsequently leveraged as the key and value inputs in the Reference-Attention layer of the corresponding Transformer block within the Denoising U-Net. The Reference U-Net's sole function is to encode the reference image, ensuring no noise is introduced and only a solitary forward pass is executed during the diffusion process. Additionally, to prevent the introduction of extraneous information, an empty text placeholder is fed into the cross-attention layer of the ReferenceNet. This meticulous design ensures accurate capture and seamless integration of the reference image's essence into the generative process, facilitating the creation of high-fidelity outputs.

\textbf{Audio Encoder.} The animation of the synthesized character is primarily driven by the nuances in pronunciation and tonality within speech. We derive the audio representation embedding for the corresponding frame by concatenating features extracted from the input audio sequence through the various processing blocks of the pre-trained Wav2Vec model\cite{schneider2019wav2vec}. The motion of the character can be influenced by both future and past audio segments, necessitating the consideration of temporal context. To address this, we define the audio features for each generated frame by concatenating the features of adjacent frames. Subsequently, we employ Audio-Attention layers in the Denoising U-Net to implement a cross-attention mechanism between the latent code and the output following each Reference-Attention layer, effectively integrating voice features into the generation procedure. This ensures that the motion of the synthesized character is finely tuned to the dynamic subtleties of the accompanying audio, consequently enhancing the realism and expressiveness of the output.

\textbf{Landmark Encoder.} Utilizing the robust spatial correspondence between each facial landmark image and its associated target frame, we integrate a Landmark Encoder into our EchoMimic framework. The Landmark Encoder, instantiated as a streamlined convolutional model, is responsible for encoding each facial landmark image into a feature representation aligned with the dimensions of the latent space. Subsequently, the encoded facial landmark image features are directly integrated with the multi-frame latents via element-wise addition before ingestion into the Denoising U-Net. This strategy enables the seamless incorporation of precise spatial information critical for maintaining accurate anatomical structure and movement in the generative process, ultimately enhancing the fidelity and coherence of the output sequences.

\textbf{Temporal Attention Layer.} In order to generate temporally coherent video sequences, EchoMimic incorporates Temporal-Attention layers to encode the temporal dynamics inherent in video data. These layers adeptly capture the intricate dependencies between successive frames by reshaping the hidden state and applying self-attention mechanisms along the temporal axis of the frame sequence. Specifically, given a hidden state $h\in R^{b\times f \times d \times h \times w}$, where $b$, $f$, $d$, $h$, and $w$ denote the batch size, the number of frames, the feature dimension, the height, and the width. Our Temporal-Attention layers adeptly capture the intricate dependencies between successive frames. This is accomplished by first reshaping the hidden state to $h\in R^{(b\times h \times w) \times f \times d}$, thereby enabling the application of self-attention mechanisms along the temporal axis of the frame sequence. Through this process, the Temporal-Attention layers discern and learn nuanced motion patterns, ensuring smooth and harmonious transitions in the synthesized frames. As a result, the video sequences demonstrate a high degree of temporal consistency, reflecting natural and fluid motion and enhancing the visual quality and realism of the generated content.

\textbf{Spatial Loss.} Since the resolution of latent space ($64*64$ for $512*512$ image) is relative too low to capture the subtle facial details, a timestep-aware spatial loss is proposed to learning the face structure directly in the pixel space. In particular, predicted latent $z_{t}$ is first mapped to $z_{0}$ by sampler. Then the predicted image is obtained via passing $z_{0}$ to the vae decoder. Finally, the mse loss is computed on the predicted image and its corresponding ground truth. Besides mse loss, LPIPS loss is adopted to further refine the details of the image. Further, since it is difficult for the model to converge when timestep $t$ is large, we propose a timestep-aware function to reduce the weight for large $t$. Detailed objective function is shown below:

\begin{equation}
\begin{aligned}
\label{equation:obj}
Obj = L_{latent} + \lambda L_{spatial}
\end{aligned}
\end{equation}

\begin{equation}
\begin{aligned}
\label{equation:obj}
L_{spatial} = w(t) [L2(I_{p}, I_{GT}) + LPIPS(I_{p}, I_{GT})]
\end{aligned}
\end{equation}

\begin{equation}
\begin{aligned}
\label{equation:obj}
w(t) = cosine(t* \pi / 2T)
\end{aligned}
\end{equation}

\subsection{Training Details}\label{section:train}
We adopt a two-stage training strategy following previous works. And we propose efficient techniques including random landmark selection and audio augmentation to boost the training process.

\textbf{Stage1.} In stage1, reference unet and denoising unet are training on single frame data to learning the relations between image-audio and image-pose. In particular, temporal attention layer is not inserted to the denoising unet in stage1.

\textbf{Stage2.} In stage2, temporal attention layer is inserted into denoising unet. And the overall pipeline is trained on 12-frame videos for the final video generation. Only the temporal model is trained while the other parts are frozen during stage2.

\textbf{Random Landmark Selection.} To achieve robust landmark-based image driven, we propose a technique called Random Landmark Selection (RLS). In particular, the face is split into several parts including eyebrows, eyes, pupils, nose and mouth. During training, we randomly drops one or several parts of the face.

\textbf{Spatial Loss and Audio Augmentation.} During our experiments, we find that two key techniques can significant improve the quality of the generated video. One is the above proposed spatial loss, which forces the diffusion model to learn the spatial information directly from the pixel space. The other is audio augmentation, which inserts noise and other perturbations to the original audios to achieve similar data augmentations as the images do.  

\subsection{Inference}\label{section:inf}
For the audio-driven case, the inference process is straightforward. While for the pose-driven or audio+pose-driven case, it is important to align the pose with the reference image according to previous works. Despite there are several techniques proposed for motion alignment, challenges still exist. For instance, existing methods usually apply full face perspective warp affine while ignoring the matching of facial parts. To this end, we propose a developed version of motion alignment called part-aware motion synchronization.

\textbf{Part-aware Motion Synchronization.} Part-aware Motion Synchronization splits the face into several parts. Then, a transformation matrix is first computed on full face. Finally, an extra residual transformation matrix is computed on each part, which will add to the previous matrix to obtain the final matrix.

\section{Experiments}
\subsection{Experimental Setups}
\textbf{Implementation Details.} The study involved experiments that entailed both the training and inference phases, conducted on a high-performance computing setup equipped with 8 NVIDIA A100 GPUs. The training process was divided into two segments, each consisting of 30,000 steps. These steps were executed with a batch size of 4, working with video data formatted at a resolution of 512 × 512 pixels. In the second phase of training, 14 video frames were generated per iteration, integrating the derived latent variables from the motion module with the initial 2 actual video frames to ensure narrative consistency. A consistent learning rate of 1e-5 was maintained throughout both training phases. The motion module was initialized utilizing pre-trained weights from the Animatediff model to expedite the learning process. To introduce variability and improve the model's generative capacity, elements including the reference image, guiding audio, and motion frames were randomly omitted with a $5\%$ chance during the training routine. For the inference stage, the system upheld sequential coherence by merging the latent variables, which had been perturbed with noise, with the feature representations extracted from the latest 2 motion frames of the preceding step within the motion module. This strategy guaranteed a seamless transition between successive video sequences, thereby enhancing the overall quality and continuity of the generated videos.

\textbf{Datasets.} We collected approximately 540 hours (each video segment is about 15 seconds long, totaling approximately 130,000 video clips) of talking head videos from the internet and supplemented this with the HDTF\cite{zhang2021flow} and CelebV-HQ\cite{zhu2022celebv} datasets to train our models. To uphold rigorous standards for training data, we implemented a meticulous data cleaning procedure. This process centered on preserving videos that feature a single person speaking, with a strong correlation between lip movements and accompanying audio, while discarding those with scene transitions, pronounced camera movements, overly expressive facial actions, or viewpoints that are fully profile-oriented. We apply the MediaPipe\cite{lugaresi2019mediapipe} to extract facial landmarks of training videos.

\textbf{Evaluation Metric.} The metrics used to assess the performance of the portrait image animation method encompass FID (Fréchet Inception Distance), FVD (Fréchet Video Distance), SSIM (Structural Similarity Index Measure) and E-FID (Expression-FID). FID and FVD gauge how closely synthetic images resemble actual data, lower scores here signify greater realism and superior performance. The SSIM indices measure the structural similarity between ground truth videos and generated videos. Additionally, E-FID leverages the Inception network's features to critically evaluate the authenticity of the generated images, providing a more nuanced gauge of image fidelity. First, E-FID employs face reconstruction method, as detailed in the\cite{deng2019accurate}, to extract expression parameters. Then, it calculates the FID of these extracted parameters to quantitatively assess the disparity between the facial expressions present in the generated videos and those found in the GT dataset.

\textbf{Baseline.} Within our quantitative experimental framework, we undertook a comparative assessment pitting our proposed methodology against several open-source implementations, namely SadTalker\cite{zhang2023sadtalker}, AniPortrait\cite{wei2024aniportrait}, V-Express\cite{wang2024v} and Hallo\cite{xu2024hallo}. This evaluation spanned datasets including HDTF, CelebV-HQ, and our collected dataset. To ensure a rigorous examination, a standard 90:10 ratio was adopted for splitting identity data, with $90\%$ dedicated to the training phase. The qualitative comparison involved evaluating our method against these selected approaches, taking into account the reference images, audio inputs, and the resulting animated outputs provided by each respective method. This qualitative assessment was aimed at gaining insights into the performance and capabilities of our method in generating realistic and expressive talking head animations.

\subsection{Quantitative Results}
\textbf{Comparison on HDTF dataset.} Table \ref{tab:metrics_comparison_hdtf} provides a quantitative evaluation of diverse portrait animation methods, focusing on the HDTF dataset. Our proposed EchoMimic excels across multiple evaluative indices. Specially, it achieves the best scores in FID at 29.136 and FVD at 492.784. These metrics affirm the heightened visual quality and high quality temporal coherence intrinsic to the generated talking head animations. Furthermore, our method's ability in lip synchronization is prominently highlighted by exceptional scores on SSIM at 0.812 and E-FID at 1.112. These evaluation results accentuate the efficacy of our method in harmoniously integrating precise lip movement synchronization with visually compelling and temporally aligned content generation.

\begin{table}[htbp]
\centering
\caption{The quantitative comparisons with the existed portrait image animation approaches on the HDTF.}
\label{tab:metrics_comparison_hdtf} 
\begin{tabular}{lllll} 
\toprule
\textbf{Method} & {\textbf{FID$\downarrow$}} & {\textbf{FVD$\downarrow$}} & {\textbf{SSIM$\uparrow$}} & {\textbf{E-FID$\downarrow$}} \\
\midrule
SadTalker\cite{zhang2023sadtalker} & 41.535 & 1138.056 & 0.790 & 2.248 \\
AniPortrait\cite{wei2024aniportrait} & {53.143} & {1038.239} & {0.751} & {1.939} \\
V-Express\cite{wang2024v} & 58.230 & 1184.203 & 0.724 & 1.807 \\
Hallo\cite{xu2024hallo} & 37.659 & 501.074 & 0.781 & 1.525 \\
\textbf{EchoMimic} & \textbf{29.136} & \textbf{492.784} & \textbf{0.812} & \textbf{1.112} \\
\bottomrule
\end{tabular}
\end{table}

\begin{figure*}
  \centering
  \includegraphics[width=0.88\linewidth]{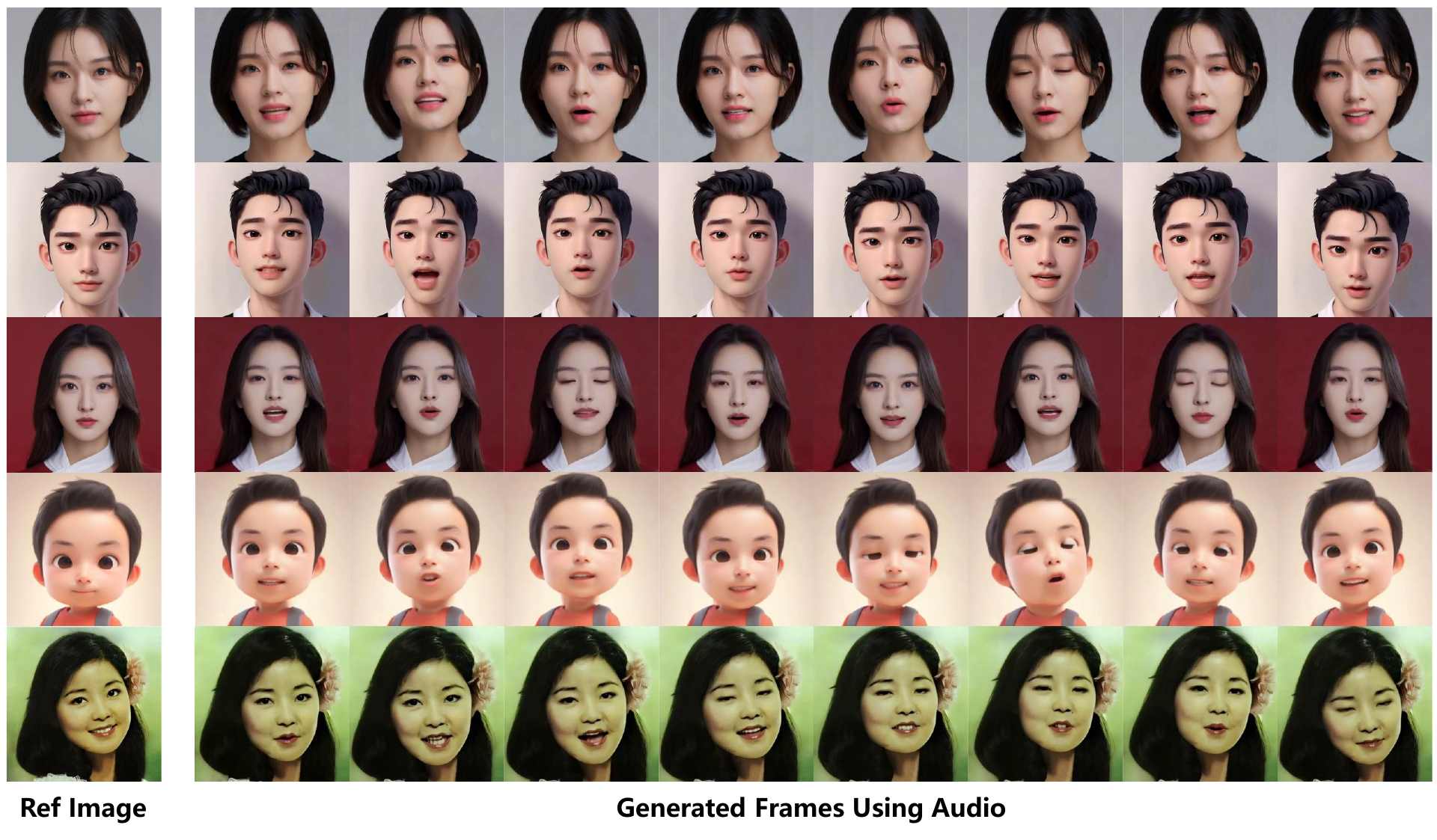}
  \caption{Video generation results of the proposed EchoMimic given different portrait styles and audios.}
  \label{fig:res_1} %% label for entire figure
\end{figure*}

\textbf{Comparison on CelebV-HQ dataset.} Compared to the HDTF and our collected datasets, the video quality of the CelebV-HQ dataset is relatively lower, consequently, all algorithms perform with lower scores on this dataset compared to the other two dataset results. The results detailed in Table \ref{tab:metrics_comparison_celebvhq} present a quantitative evaluation of various portrait animation methods employing the CelebV-HQ dataset. Our proposed EchoMimic also achieves comparable results on this dataset compared with other methods. Specifically, it attains the lowest values in FID at 63.258. Notably, our method yields the most favorable E-FID score of 2.723 among the compared techniques, highlighting its capacity to generate high-fidelity animations that maintain a striking level of temporal consistency.

\begin{table}[htbp]
\centering
\caption{The quantitative comparisons with the existed portrait image animation approaches on the CelebV-HQ dataset.}
\label{tab:metrics_comparison_celebvhq} 
\begin{tabular}{lllll} 
\toprule
\textbf{Method} & {\textbf{FID$\downarrow$}} & {\textbf{FVD$\downarrow$}} & {\textbf{SSIM$\uparrow$}} & {\textbf{E-FID$\downarrow$}} \\
\midrule
SadTalker\cite{zhang2023sadtalker} & 93.883 & 1454.328 & 0.641 & 3.971 \\
AniPortrait\cite{wei2024aniportrait} & {92.003} & {1297.805} & {0.609} & {3.916} \\
V-Express\cite{wang2024v} & 95.483 & 2126.248 & 0.524 & 4.720 \\
Hallo\cite{xu2024hallo} & 70.420 & \textbf{1073.718} & \textbf{0.644} & 2.851 \\
\textbf{EchoMimic} & \textbf{63.258} & 1115.857 & 0.633 & \textbf{2.723} \\
\bottomrule
\end{tabular}
\end{table}

\textbf{Comparison on our collected dataset.} We also evaluate proposed EchoMimic and other portrait animation methods on our collected datasets. Table \ref{tab:metrics_comparison_collect} shows the evaluation results. Our proposed EchoMimic exhibits the lowest scores in both FID at 43.272 and FVD at 988.144, signifying a marked improvement in visual quality and temporal consistency over existing techniques. Besides, our method achieves comparable SSIM score (0.691) to the best-performing algorithm (0.699). Furthermore, it also records the most favorable E-FID score of 1.421, reinforcing its capability to generate animations of superior fidelity even under challenging and diverse scenarios. These quantitative findings collectively emphasize the robustness and efficacy of our technique, affirming its aptitude for creating high-quality animations that are not only temporally coherent but also excel in precise lip synchronization.

\begin{table}[htbp]
\centering
\caption{The quantitative comparisons with the existed portrait image animation approaches on the our collected dataset.}
\label{tab:metrics_comparison_collect} 
\begin{tabular}{lllll} 
\toprule
\textbf{Method} & {\textbf{FID$\downarrow$}} & {\textbf{FVD$\downarrow$}} & {\textbf{SSIM$\uparrow$}} & {\textbf{E-FID$\downarrow$}} \\
\midrule
SadTalker\cite{zhang2023sadtalker} & 64.633 & 1681.836 & \textbf{0.699} & 2.150 \\
AniPortrait\cite{wei2024aniportrait} & {66.884} & {2054.527} & {0.665} & {2.312} \\
V-Express\cite{wang2024v} & 62.721 & 2103.213 & 0.658 & 1.689 \\
Hallo\cite{xu2024hallo} & 50.474 & 1405.215 & 0.690 & 1.452 \\
\textbf{EchoMimic} & \textbf{43.272} & \textbf{988.144} & 0.691 & \textbf{1.421} \\
\bottomrule
\end{tabular}
\end{table}

\begin{figure*}
  \centering
  \includegraphics[width=0.88\linewidth]{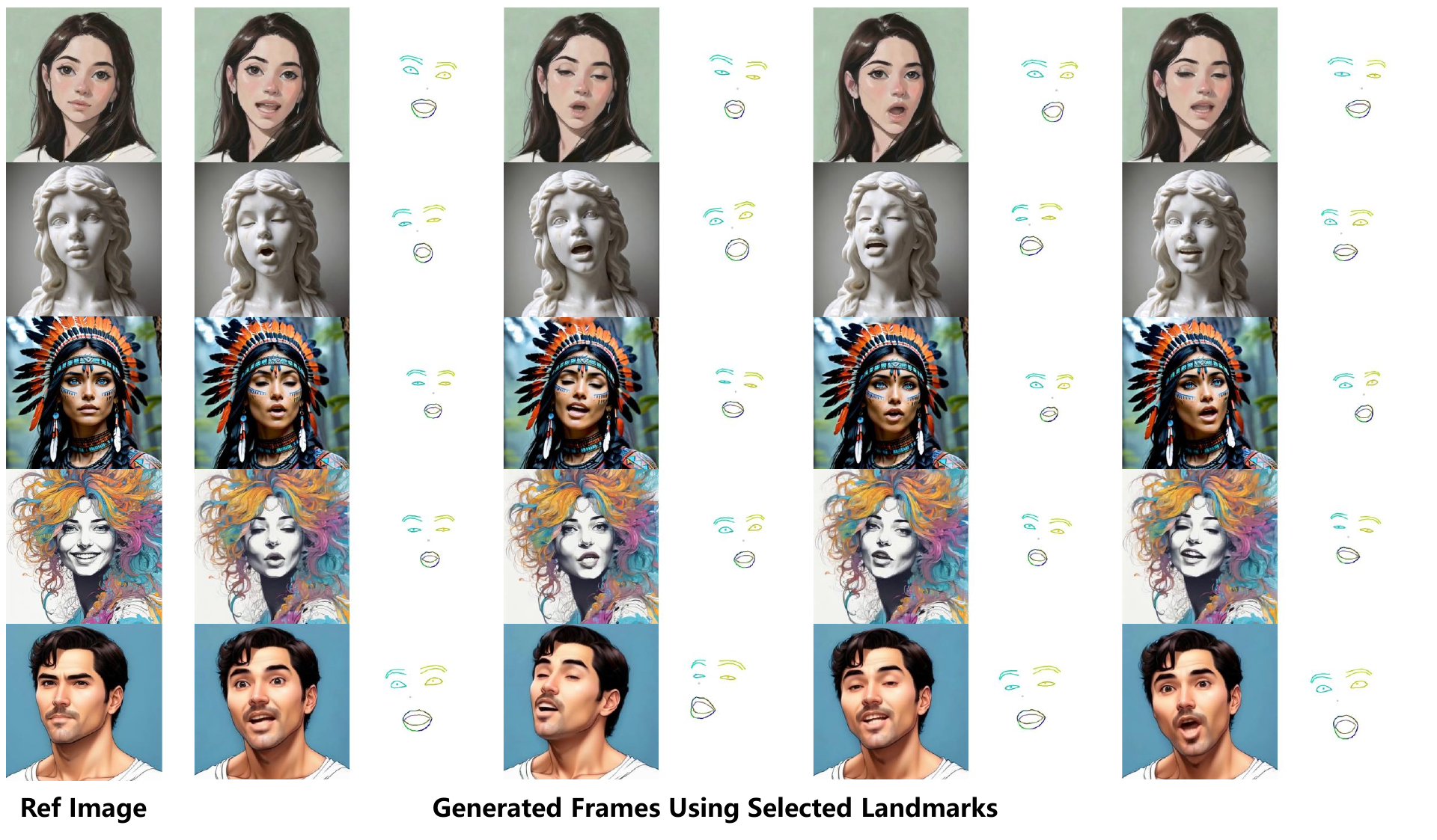}
  \caption{Video generation results of the proposed EchoMimic given different portrait styles and landmarks.}
  \label{fig:res_2} %% label for entire figure
\end{figure*}

\subsection{Qualitative Results}
As mentioned in previous sections, we have proposed three driving modes, which are audio driven, landmark driven and audio+selected landmark driven. In this section, we will qualitatively evaluate the driving effects of these three modes and further present the generation results of EchoMimic across various portrait styles. The corresponding video results can be accessed via our project homepage.

\textbf{Audio Driven.} Audio driven refers to generating a talking head video utilizing solely the input audio signal and a reference image. Fig. \ref{fig:res_1} shows the qualitative results with various portrait styles and different input audios. The obtained results show the adaptability and resilience of the proposed approach in synthesizing a broad spectrum of audiovisual outputs. These results also show that our approach is proficient in managing diverse audio inputs to yield high-resolution, visually consistent talking head videos that exhibit seamless synchronization with the accompanying audio content. These results also underscore the efficacy and robustness of our approach across a myriad of acoustic scenarios, thereby affirming its potential for advancing the state-of-the-art in audio-driven video generation.

\textbf{Landmark Driven.} Landmark driven means to generating a talking head video utilizing a reference image and landmark controls, which achieves the same functionality as described in Follow Your Emoji\cite{ma2024follow}. Fig. \ref{fig:res_2} shows landmark mapping results with motion synchronization method. We can see from the results compared with previous facial landmark mapping method, with our proposed motion synchronization the landmarks from driving frames are well aligned with reference image, which facilitates the algorithm to generate control results that are more akin to the face shape in the reference image. For instance, in the mapping of the potato man's face depicted, the algorithm is capable of projecting the small mouth from the reference image onto the significantly larger mouth of the potato man. Fig. \ref{fig:res_2} shows the qualitative results with various portrait styles with different landmarks. It can be seen that our proposed method performs better in expression transfer and preserves the identity of reference portraits during animation at the same time. Besides, our proposed method also demonstrates an enhanced capability in addressing substantial variations in pose and accurately reproducing nuanced expressions.

\begin{figure*}
  \centering
  \includegraphics[width=0.88\linewidth]{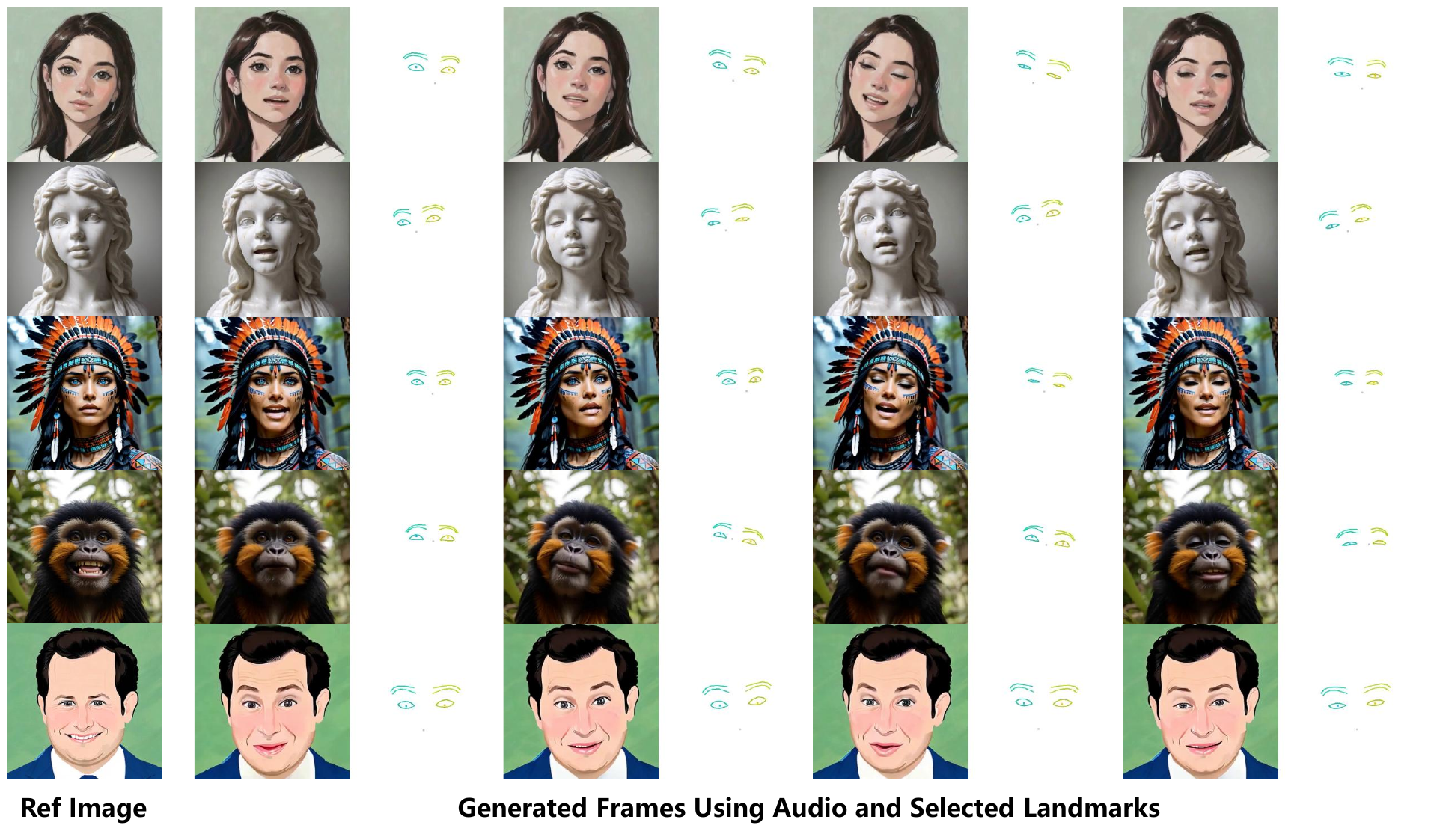}
  \caption{Video generation results of the proposed EchoMimic given different portrait styles, audios and selected landmarks.}
  \label{fig:res_3} %% label for entire figure
\end{figure*}

\textbf{Audio + Selected Landmark Driven.} Audio + selected landmark driven means to generating a talking head video utilizing the input audio signal along with a reference image and selected landmark controls. Using this kind of driven mode, we can not only maintain naturally lip synchronization, but also allow for finer control over facial details. Besides, this driven mode enables the generated video to exhibit the facial expressions and actions desired by the user, such as blinking or closing eyes while singing, further enhancing the authenticity of the generated video. Fig. \ref{fig:res_3} shows the qualitative results with various portrait styles with different input audios and selected landmarks. Similar with previous results, this driven mode also yields clear generated videos while ensuring lip movements synchronized with the audio. Furthermore, as show from results, the generated videos are precisely controlled by our selected landmarks, which enables the generation of facial expressions that are consistent with pre-defined landmarks. 

\subsection{Ablation Study}
\begin{figure*}
  \centering
  \includegraphics[width=0.88\linewidth]{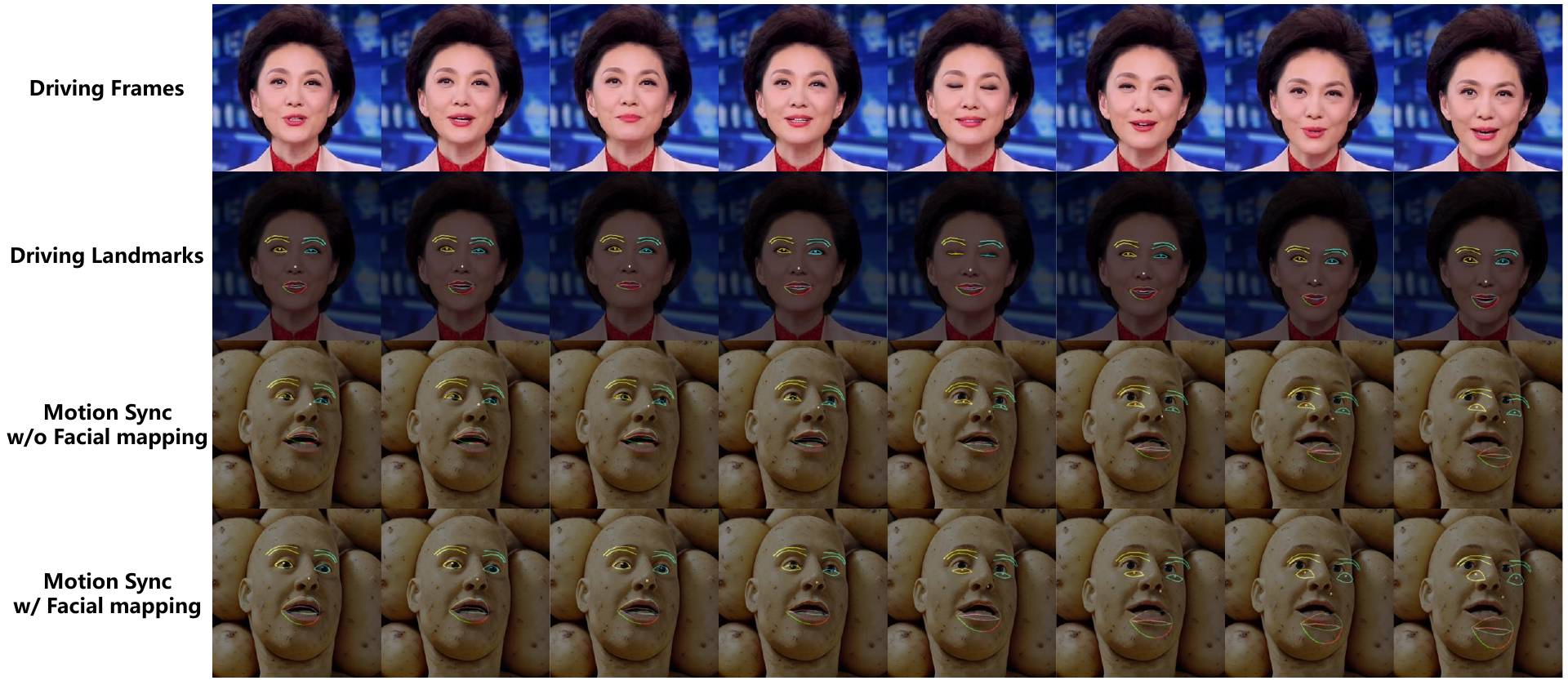}
  \caption{Landmark mapping results with motion synchronization method.}
  \label{fig:ab} %% label for entire figure
\end{figure*}

\textbf{Facial Landmark Mapping with Motion Synchronization.} In this ablation study, we aim to validate the efficacy of our proposed motion synchronization method. Landmark driven means to generating a talking head video utilizing a reference image and landmark controls, which achieves the same functionality as described in Follow Your Emoji. Fig. \ref{fig:ab} shows landmark mapping results with motion synchronization method. We can see from the results compared with previous facial landmark mapping method, with our proposed motion synchronization the landmarks from driving frames are well aligned with reference image, which facilitates the algorithm to generate control results that are more akin to the face shape in the reference image. For instance, in the mapping of the potato man's face depicted, the algorithm is capable of projecting the small mouth from the reference image onto the significantly larger mouth of the potato man. 

\textbf{Facial Expression Control by Selected Landmarks.} In this ablation study, we evaluate the effects of three different driving modes: (1) audio driven, (2) facial landmark driven, and (3) audio with selected facial landmark driven. Table \ref{tab:ab_mode} provides a quantitative evaluation of diverse portrait animation methods, focusing on the HDTF dataset. The experimental findings demonstrate that the first driving mode, characterized by its weakest constraints, affords increased freedom in the generated videos, leading to larger disparities from the original videos. The second driving mode, characterized by its strong facial landmark constraints, results in the closest resemblance between the generated videos and the original videos, thus delivering the most favorable results. The third driving mode serves as a compromise between the first and second modes, yielding intermediate outcomes in both freedom and similarity.

\begin{table}[htbp]
\centering
\caption{The quantitative comparisons with different driving modes of EchoMimic on HDTF dataset. ``A'' represents the audio only driving model, ``L'' represents the pose only driving model, and ``A+L'' refers to the mode where the video is generated based on both audio and landmarks without mouth region.}
\label{tab:ab_mode} 
\begin{tabular}{cllll} 
\toprule
\textbf{Driving Mode} & {\textbf{FID$\downarrow$}} & {\textbf{FVD$\downarrow$}} & {\textbf{SSIM$\uparrow$}} & {\textbf{E-FID$\downarrow$}} \\
\midrule
A & 29.136 & 492.784 & 0.812 & 1.112 \\
L & 22.970 & 156.537 & 0.889 & 1.057 \\
A+L & 22.981 & 181.741 & 0.885 & 1.093 \\
\bottomrule
\end{tabular}
\end{table}

\subsection{Limitation and Future Work}
While this study introduces notable advancements in the realm of portrait image animation, it is imperative to acknowledge that there still exists several inherent limitations that need further investigation and improvement. Future research endeavors are poised to refine and expand upon the presented methodology, thereby contributing to its advancement and enhancement. \textbf{(1) Update to video processing frameworks.} The current architecture, while demonstrating fairly decent performance in handling video content, is essentially an extension of Stable Diffusion image processing techniques applied to the video domain. Consequently, this framework does not qualify as a genuine video processing framework in the true sense. Future work could explore leveraging authentic video processing frameworks (such as 3DVAE, DiT etc.) to reformulate this method, thereby facilitating enhancements and optimizations tailored specifically for video content\cite{peebles2023scalable,yu2023language}. \textbf{(2) Use acceleration technique.} Currently, there is a proliferation of algorithms emerging that accelerate Stable Diffusion generation process\cite{chai2023speedupnet,luo2023latent}, and subsequent research can harness these algorithms to speed up EchoMimic framework, thereby achieving real-time generation capabilities. This real-time generation can be broadly used for applications in scenarios such as real-time digital human interactions and conversations.

\section{Conclusions}
In this paper, we present EchoMimic, a novel portrait image animation approach that leverages audio signals and facial landmarks to generate high-quality and expressive talking head videos. Through a novel training strategy, EchoMimic achieves significant advancements in generating authentic and visually appealing portrait animations. The comprehensive evaluations conducted across diverse public datasets and meticulous comparison with alternative algorithms underscore the superior performance and robustness of EchoMimic. By addressing key challenges in portrait animation, our approach showcases substantial promise for enhancing multimedia experiences and advancing the state-of-the-art in video synthesis. The detailed methodology, qualitative and quantitative assessments, and ablation studies collectively reinforce the efficacy and potential impact of EchoMimic in the field of portrait image animation.

%%%%%%%%% REFERENCES
{\small
\bibliographystyle{ieee_fullname}
\bibliography{egbib}
}
% \end{multicols}
\end{document}